\documentclass[sigconf]{acmart}
\AtBeginDocument{%
  }

\setcopyright{acmlicensed}
\copyrightyear{2018}
\acmYear{2018}
\acmDOI{XXXXXXX.XXXXXXX}
\acmConference[Conference acronym 'XX]{Make sure to enter the correct
  conference title from your rights confirmation email}{June 03--05,
  2018}{Woodstock, NY}

\acmISBN{978-1-4503-XXXX-X/2018/06}

\renewcommand\footnotetextcopyrightpermission[1]{}

\usepackage{graphicx}
\def\equationautorefname~#1\null{%
  Equation~(#1)\null
}
\usepackage[most]{tcolorbox}
\usepackage{algorithm}
\usepackage[noend]{algpseudocode}
\usepackage{xcolor}
\usepackage{xspace}
\usepackage{diagbox}
\usepackage{pifont}
\usepackage{hyperref}
\usepackage{subcaption}

\usepackage[table,xcdraw]{xcolor}
\usepackage{colortbl}

\usepackage{footmisc}
\usepackage{makecell} 

\usepackage[colorinlistoftodos]{todonotes}
\usepackage{booktabs}
\usepackage{upgreek}
\usepackage{textcomp}
\usepackage{bbding}
\usepackage{graphics}
\usepackage{enumitem}
\usepackage{amsmath}

\setlist[itemize]{%
  leftmargin=15pt,
  topsep=3pt,
  itemsep=1pt,
  parsep=0pt,
  partopsep=0pt
}
\setlist[enumerate]{%
  leftmargin=15pt,
  topsep=3pt,
  itemsep=1pt,
  parsep=0pt,
  partopsep=0pt
}

\usepackage{multirow}
\usepackage{float}
\usepackage{amsmath,bm}
\usepackage{color}

\usepackage{mathrsfs}
\usepackage[T1]{fontenc}

\definecolor{uurl}{HTML}{000099}

\AtEndPreamble{
	\usepackage{hyperref}

        \def\equationautorefname~#1\null{Equation~(#1)\null}
        
	\hypersetup{
		colorlinks = true,
		linkcolor = purple,
		anchorcolor = purple,
		citecolor = purple,
		filecolor = purple,
		urlcolor = purple
	}
}

\newcommand{\toolname}{QMFOL\xspace}
\newcommand{\benchmarkname}{QMFOLBench\xspace}
\newcommand{\task}{\ensuremath{\mathcal{T}}}
\newcommand{\premises}{\ensuremath{\mathcal{P}}}
\newcommand{\conclusion}{\ensuremath{\mathcal{C}}}
\newcommand{\correctnesslabel}{\ensuremath{\mathcal{L}}}
\newcommand{\rules}{\ensuremath{\mathcal{R}}}
\newcommand{\facts}{\ensuremath{\mathcal{F}}}
\newcommand{\nodes}{\ensuremath{\mathcal{N}}}
\newcommand{\true}{\textit{True}}
\newcommand{\false}{\textit{False}}
\newcommand{\unknown}{\textit{Unknown}}

\definecolor{mblue}{HTML}{0066CC} 
\definecolor{mpink}{HTML}{FF66B3} 
\definecolor{mGray}{rgb}{0.5,0.5,0.5}
\newcommand{\commentstyle}[1]{\textcolor{mGray}{\small{#1}}}

\definecolor{mRed}{HTML}{d73027}
\definecolor{mGreen}{HTML}{1a9850}
\definecolor{mOrange}{HTML}{fdae61}

\definecolor{TopicColor}{HTML}{E07A7A}
\definecolor{DepthColor}{HTML}{8074C8}
\definecolor{WidthColor}{HTML}{7895C1}
\definecolor{LabelColor}{HTML}{82AD7F}
\definecolor{DistractorColor}{HTML}{E6A96F}

\newcommand{\arrowcomp}[3]{%
  \ensuremath{%
    \ifdim #1pt<#2pt
      #2_{\textcolor{mRed}{\uparrow#3}}%
    \else
      \ifdim #1pt>#2pt
        #2_{\textcolor{mGreen}{\downarrow#3}}%
      \else
        #2%
      \fi
    \fi
  }%
}

\usepackage{tcolorbox}
\usepackage{tikz}

\usepackage{svg}
\begin{document}


\title{QMFOL: Benchmarking Large Language Model Reasoning via Quantifiable Monadic First-Order Logic Test Case Generation}
\author[X Zheng]{Xinyi Zheng}
\email{xinyizheng@hust.edu.cn}
\affiliation{%
  \institution{Huazhong University of Science and Technology}
  \city{Wuhan}           
  \country{China}
}

\author[L Shi]{Ling Shi}
\email{ling.shi@ntu.edu.sg}
\affiliation{%
  \institution{Nanyang Technological University}
  \city{Singapore}           
  \country{Singapore}
}

\author[T Yu]{Tianlong Yu}
\email{tommyyu21@163.com}
\affiliation{%
  \institution{Hubei University}
  \city{Wuhan}           
  \country{China}
}

\author[Y Zhao]{Yongxin Zhao}
\email{yxzhao@sei.ecnu.edu.cn}
\affiliation{%
  \institution{East China Normal University}
  \city{Shanghai}           
  \country{China}
}

\author[L Goette]{Lorenz Goette}
\email{lorenz@lorenzgoette.org}
\affiliation{%
  \institution{National University of Singapore}
  \city{Singapore}           
  \country{Singapore}
}

\author[K Wang]{Kailong Wang}
\authornote{Kailong Wang~(wangkl@hust.edu.cn) is the corresponding authors.}
\email{wangkl@hust.edu.cn}
\affiliation{%
  \institution{Huazhong University of Science and Technology}
  \city{Wuhan}           
  \country{China}
}

\renewcommand{\shortauthors}{Zheng et al.}

\begin{abstract}

Large Language Models (LLMs) have made significant progress in reasoning, particularly in deductive reasoning, which is crucial for high-stakes decision-making. As models improve, evaluation benchmarks should evolve to keep pace. However, existing benchmarks lack fine-grained control over logical complexity and struggle to balance semantic diversity with logical consistency.

To address these issues, we propose \toolname, an automated framework for generating monadic first-order logic reasoning tasks with quantifiable and controllable complexity. It constructs formal logical structures using conjunction and disjunction patterns, enabling precise control over reasoning depth, width, label types, and distractors. These structures are then translated into natural language via LLMs, with logical consistency ensured through round-trip verification using an external prover.
Based on our framework, we build \benchmarkname, a benchmark comprising 2880 instances with 960 configurations across diverse logical and semantic dimensions. Evaluations on six large reasoning models (LRMs) and two LLMs show that performance degrades and computational overhead increases with rising logical complexity. Models perform better on True-labeled tasks than on False or Unknown ones, and exhibit sensitivity to semantic variation.
Overall, \toolname offers a scalable and reliable approach for constructing deductive reasoning benchmarks with controllable complexity, enabling more precise evaluation of reasoning capabilities in modern language models.

\end{abstract}

\maketitle

\section{Introduction}

Large Language Models (LLMs) have made significant progress in natural language understanding, generation, and reasoning. Among these abilities, reasoning advances most rapidly. 
Recent Large Reasoning Models (LRMs) are released and updated at an unprecedented pace~\cite{deepseekai2025deepseekv32,gpt54,gemini31pro,claudesonnet46}. 
As model capabilities improve, evaluation methods must also advance to ensure that new versions achieve real gains rather than regressions or marginal improvements. This trend highlights the need for rigorous evaluation~\cite{11205366,Mohammadi_2025}.
Prior work develops dynamic benchmarks spanning mathematics \cite{huang2025mathperturb,mirzadeh2024gsm}, commonsense reasoning \cite{li2024drowzee}, temporal reasoning \cite{islakoglu2025chronosense,chu2024timebench}, and logical reasoning \cite{qi2025proverqa,xu2025socrates}. These benchmarks evaluate reasoning across diverse tasks and help mitigate data contamination in existing datasets \cite{zhao-etal-2025-coreeval,choi2025contaminatedbenchmarkquantifyingdataset}.
In particular among them all, deductive reasoning \cite{pan-etal-2023-logic,nguyen2025noninteractivesymbolicaidedchainofthoughtlogical} is central to decision making \cite{eigner2024determinantsllmassisteddecisionmaking,yang2023foundation}. It requires deriving conclusions from explicit premises and supports systematic and interpretable reasoning. 
This capability is critical in high-stakes and knowledge-intensive reasoning scenarios, such as law~\cite{LAI2024181} and healthcare~\cite{Clusmann2023}. Therefore, accurately evaluating the deductive reasoning ability of language models is essential.

Existing work proposes several deductive reasoning benchmarks, as shown in \autoref{table:benchmarks}.
RuleTaker~\cite{clark2020ruletaker}, ProofWriter~\cite{tafjord2021proofwriter}, RobustLR~\cite{sanyal2022robustlr}, and PrOntoQA-OOD~\cite{saparov2022prontoqa,saparov2023prontoqaood} generate logical structures using predefined algorithms and convert them into natural language with fixed templates. These benchmarks are relatively early and may overlap with training data, leading to data contamination. Their template-based design also limits semantic diversity.
FOLIO~\cite{han2024folio} is manually constructed based on common logical patterns. It provides richer semantics and more natural language expressions. However, it does not scale well, and human errors may introduce logical inconsistencies.
ProverQA~\cite{qi2025proverqa} expands logical structures by randomly applying logical rules and verifies them with theorem provers, then uses LLMs for natural language generation. Although this improves semantic diversity, it is unclear whether the logical structure is preserved during conversion.

Despite these efforts, existing benchmarks for deductive reasoning still face key challenges.
\textbf{Challenge \#1: Limited control over logical complexity across dimensions.} 
Most benchmarks increase overall complexity by incrementally applying reasoning rules or transformations, but lack fine-grained control over individual dimensions.
For example, varying reasoning depth while keeping width and other factors fixed remains difficult, limiting the evaluation of model performance along a single dimension.
\textbf{Challenge \#2: Trade-off between semantic diversity and logical consistency in natural language conversion.} 
Converting formal logical structures into natural language remains challenging. Template-based methods preserve logical correctness but limit semantic diversity, while manual and LLM-based methods improve linguistic richness but may introduce hallucinations, making it difficult to ensure consistency between the original logic and the generated text.

\begin{table*}[!htbp]
\centering
\small
\fontsize{9}{11}\selectfont
\caption{
The comparison of existing deductive reasoning benchmarks. 
Dimensions represent the factors considered in the benchmark.
RuleTaker and PrOntoQA-OOD consider only \true{}/\false{} labels, while the others additionally consider \unknown{}.
}
\vspace{-0.3cm}
\label{table:benchmarks}
\resizebox{\textwidth}{!}{
\begin{tabular}{lccccl}
\Xhline{1pt}
\multicolumn{1}{l}{\multirow{2}{*}{\textbf{Benchmark}}}
& \multicolumn{1}{c}{\multirow{1}{*}{\textbf{Logic}}}
& \multicolumn{1}{c}{\multirow{1}{*}{\textbf{Text}}}
& \multicolumn{1}{c}{\multirow{2}{*}{\textbf{Scalability}}}
& \multicolumn{1}{c}{\multirow{1}{*}{\textbf{Consistency}}}
& \multicolumn{1}{l}{\multirow{2}{*}{\textbf{Dimensions}}}
\\
\multicolumn{1}{c}{}
& \multicolumn{1}{c}{\multirow{1}{*}{\textbf{Construction}}}
& \multicolumn{1}{c}{\multirow{1}{*}{\textbf{Generation}}}
& \multicolumn{1}{c}{}
& \multicolumn{1}{c}{\multirow{1}{*}{\textbf{Assurance}}}
& \multicolumn{1}{c}{}
\\
\hline
RuleTaker~\cite{clark2020ruletaker} 
& Algorithm
& Template
& \textcolor{mGreen}{\ding{51}}  
& \textcolor{mGreen}{\ding{51}}  
& Depth~(0-5)
\\
ProofWriter~\cite{tafjord2021proofwriter} 
& Algorithm
& Template 
& \textcolor{mGreen}{\ding{51}} 
& \textcolor{mGreen}{\ding{51}}  
& Depth~(0-5), Label Definition~(w/o \unknown{})
\\
RobustLR~\cite{sanyal2022robustlr} 
& Algorithm
& Template
& \textcolor{mGreen}{\ding{51}} 
& \textcolor{mGreen}{\ding{51}}  
& Symbol Combination~(4), Rule Type~(6)
\\
PrOntoQA-OOD~\cite{saparov2023prontoqaood} 
& Algorithm
& Template
& \textcolor{mGreen}{\ding{51}} 
& \textcolor{mGreen}{\ding{51}}  
&  Rule Type~(6), Depth~(2-5), Width~(2-5)
\\
FOLIO~\cite{han2024folio} 
& Manual
& Manual 
& \textcolor{mRed}{\ding{55}} 
& \textcolor{mRed}{\ding{55}}  
& Depth~(0-7)
\\
ProverQA~\cite{qi2025proverqa} 
& Tool
& LLM 
& \textcolor{mGreen}{\ding{51}} 
& \textcolor{mRed}{\ding{55}}  
& Depth~(1-9), Distractor~(w/o)
\\
\hline
\multicolumn{1}{l}{\multirow{3}{*}{\textbf{\benchmarkname{}~(Ours)}}}
& \multicolumn{1}{c}{\multirow{3}{*}{Algorithm}}
& \multicolumn{1}{c}{\multirow{3}{*}{LLM}}
& \multicolumn{1}{c}{\multirow{3}{*}{\textcolor{mGreen}{\ding{51}} }}
& \multicolumn{1}{c}{\multirow{3}{*}{\textcolor{mGreen}{\ding{51}} }}
& Depth~(5,10,15,20), Width~(5,10,15,20),
\\
\multicolumn{1}{c}{}
& \multicolumn{1}{c}{}
& \multicolumn{1}{c}{}
& \multicolumn{1}{c}{}
& \multicolumn{1}{c}{}
& Distractor Number~(0,5,10,15,20),
\\
\multicolumn{1}{c}{}
& \multicolumn{1}{c}{}
& \multicolumn{1}{c}{}
& \multicolumn{1}{c}{}
& \multicolumn{1}{c}{}
& Topic~(\emph{Food},\emph{Animal},\emph{University},\emph{Mathematics})
\\
\Xhline{1pt}
\end{tabular}
}
\vspace{-4mm}
\end{table*}

\noindent \textbf{Our Work.} 
To address these challenges, we propose \toolname{}, an automated framework for generating \textbf{Q}uantifiable \textbf{M}onadic \textbf{F}irst-\textbf{O}rder \textbf{L}ogic reasoning tasks with fine-grained control over multiple dimensions. 
In essence, \toolname{} constructs formal logical expressions grounded in conjunction and disjunction patterns, and translates them into natural language via LLMs.
In the logic construction stage, we adopt Monadic First-Order Logic~(MFOL) as the formal representation, which restricts predicates to unary form while preserving core quantifier structures and improving controllability over full FOL. 
Leveraging conjunction and disjunction patterns, \toolname{} constructs logical structures with configurable depth and width, derives labeled fact–conclusion pairs, and introduces distractors to increase task difficulty~(\textbf{Challenge \#1}).
In the natural language translation stage, we use prompt-based LLM to translate MFOL instances into text based on a specific topic keyword. 
We first map predicates to domain semantics, then generate natural language descriptions from the underlying logical structures and mappings.
To ensure logical consistency, we perform round-trip verification using an external prover and repeat generation until the predicted label matches the ground truth ~\textbf{(Challenge \#2)}. 
With inputs depth, width, label type, number of distractors, and a topic keyword, \toolname{} generates controlled deductive reasoning tasks.

\noindent \textbf{Results and Findings.}
Using \toolname{}, we construct \benchmarkname{}, which spans four depth levels, four width levels, three label types, five distractors, and four topic domains, resulting in 960 unique configurations. For each configuration, we generate instances with three different random seeds, yielding 2880 instances.
Based on \benchmarkname{}, we evaluate six LRMs and two LLMs. Gemini-3.1-Pro achieves the highest accuracy, while GPT-5.4-High offers the best trade-off between performance and cost. The open-source Qwen3.5-27B performs strongly despite its smaller size, surpassing DeepSeek-V3.2-Thinking and Claude-Sonnet-4-6.
Across evaluations, performance degrades and computational overhead rises with increasing logical complexity in depth and width. Models perform better on task labeled $True$ than on $False$ and $Unknown$, with many errors stemming from insufficient reasoning. Accuracy further drops as distractors increase, although advanced models such as Gemini-3.1-Pro and GPT-5.4-High show stronger robustness. Finally, performance varies across topics even under identical logical structures, indicating semantic dependence.
Overall, the results highlight the effectiveness and necessity of our benchmark for evaluating the reasoning capabilities of LLMs and LRMs.

\noindent \textbf{Contributions.} 
The main contributions of this paper are:
\begin{itemize}[leftmargin=15pt]

    \item \textbf{A quantifiable MFOL task generation framework.}
    We propose \toolname{}~\cite{QFOL}, an automated framework that constructs formal logical structures, derives labeled reasoning tasks with controlled difficulty, and translates them into natural language with verification, enabling scalable and reliable evaluation.
    
    \item \textbf{A systematically constructed benchmark.}
    Using \toolname{}, we build \benchmarkname{}~\cite{QFOL}, a comprehensive benchmark covering 960 configurations across logical and semantic factors, and 2880 instances in total, enabling structured and scalable evaluation of model reasoning capabilities.
    
    \item \textbf{Comprehensive evaluation.} 
    We evaluate six LRMs and two LLMs on \benchmarkname{}, showing that performance degrades and overhead increases with increasing logical complexity. Models handle $True$ tasks better than $False$ or $Unknown$, and accuracy varies across topics, revealing semantic dependence.
\end{itemize}

\begin{figure}[t]
    \centering
    \includegraphics[width=0.3\textwidth]{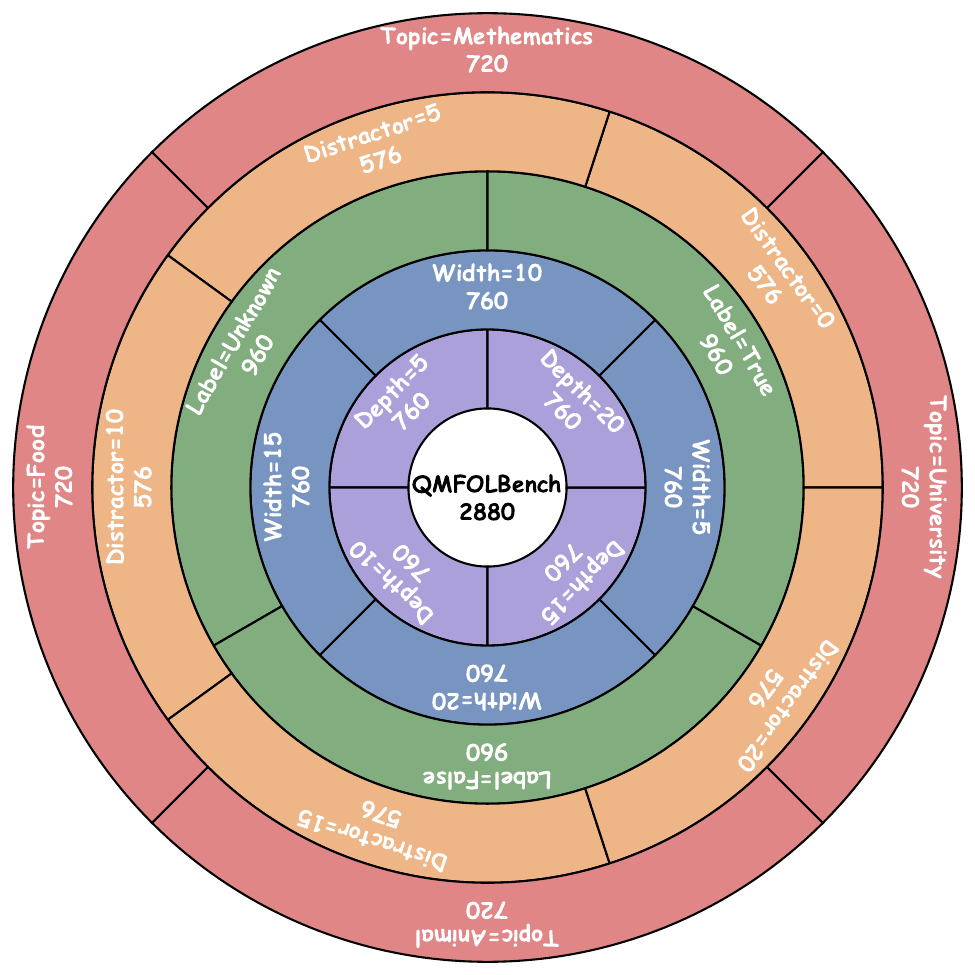}
    \vspace{-0.3cm}
    \caption{
    Data distribution of \benchmarkname{}. The benchmark spans four \textcolor{DepthColor}{depth}s, four \textcolor{WidthColor}{width}s, three \textcolor{LabelColor}{label}s, five \textcolor{DistractorColor}{distractor} levels and four \textcolor{TopicColor}{topic}s, yielding 960 configurations with 3 tasks each (2880 total). 
    }
    \vspace{-0.6cm}
    \label{fig:benchmark_distribution}
\end{figure} 

\vspace{-0.2cm}
\section{Background}
\label{sec:background}

\subsection{Deductive Reasoning Dimensions}

Although the benchmarks in \autoref{table:benchmarks} are not explicitly dimension-driven, they incorporate various analytical factors.
Reasoning depth is considered by most benchmarks, with varying definitions and typically remaining below 10.
ProofWriter~\cite{tafjord2021proofwriter} employs two labeling schemes, assigning unstated facts to either $False$ or $Unknown$.
RobustLR~\cite{sanyal2022robustlr} explores four logical operator combinations (\texttt{NOT}, \texttt{NOT+AND}, \texttt{OR+NOT}, \texttt{NOT+AND+OR}) and introduces three perturbation and three equivalence transformation rules.
PrOntoQA-OOD~\cite{saparov2023prontoqaood} considers reasoning width and six deduction rules.
ProverQA~\cite{qi2025proverqa} compares settings with and without distractors to assess their impact.
These benchmarks highlight that logical reasoning tasks involve multiple dimensions. Beyond overall performance, evaluating individual dimensions is essential for identifying and improving specific weaknesses in model reasoning ability.

\subsection{Preliminaries}
\label{sec:preliminaries}
We outline the knowledge foundations and formal definitions underlying our work.

\begin{figure*}[t]
\vspace{-0.4cm}
    \centering
    \includegraphics[width=0.96\textwidth]{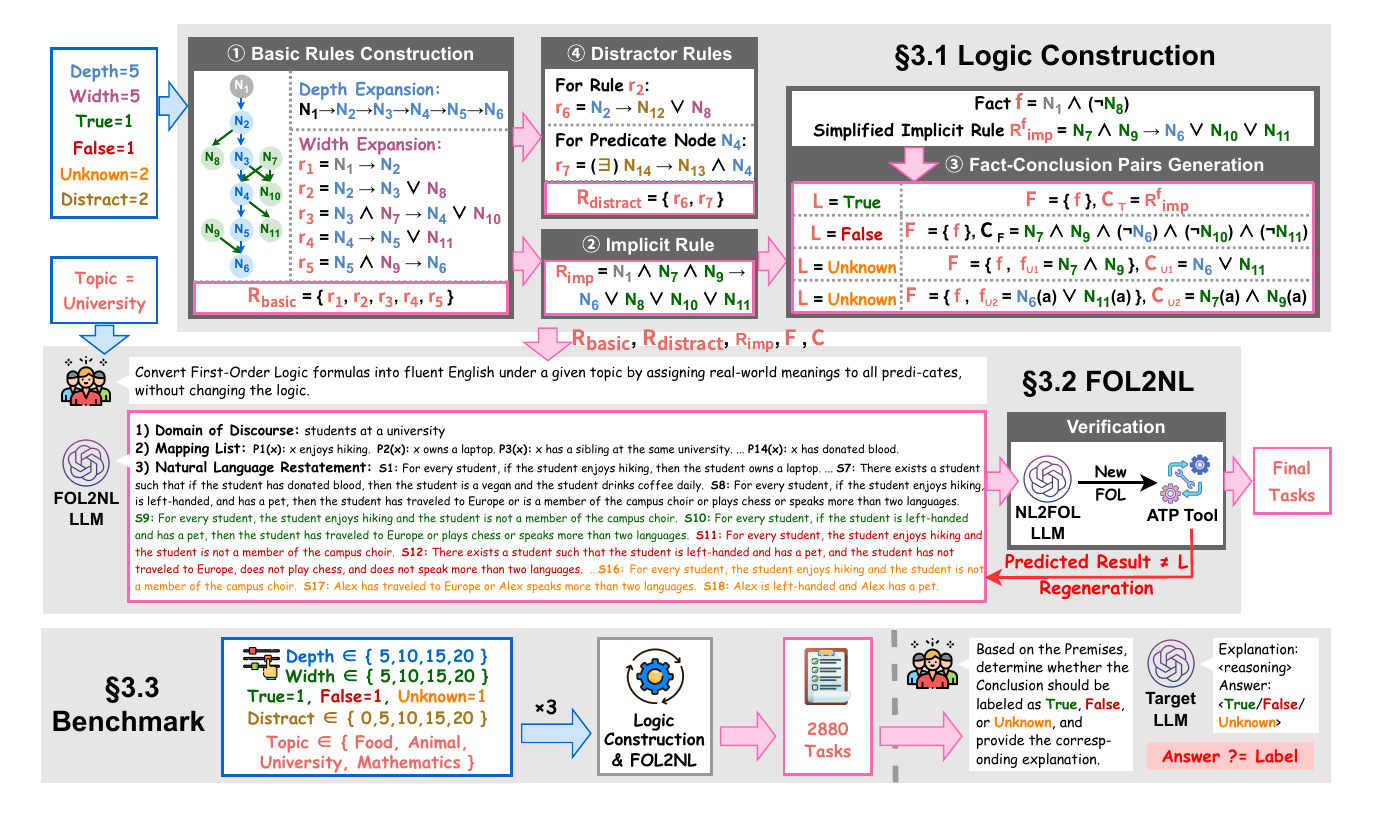}\vspace{-0.5cm}
    \caption{Overview of \toolname{}. \textcolor{mblue}{Blue} boxes represent inputs, while \textcolor{mpink}{pink} represent outputs. $N_{i}$ is a predicate node, either an atomic predicate $P_{i}$ or its negation $\neg P_{i}$. For simplicity, $N_{i} = P_{i}$ in the shown task instance.}\vspace{-0.3cm}
    \label{fig:overview}
\end{figure*}

\noindent \textbf{First-Order Logic.}
FOL is a formal language for representing and reasoning about logical statements, widely used in deductive reasoning.
It extends propositional logic with predicates and quantifiers, enabling expressions about objects in a domain. 
The core components of FOL include: \emph{constants} representing specific objects, \emph{predicates} representing properties or relations, \emph{logical connectives} such as negation ($\neg$), conjunction ($\wedge$), disjunction ($\lor$), and implication ($\rightarrow$), universal quantifier~($\forall$) and existential quantifier~($\exists$).

\noindent \textbf{Monadic First-Order Logic.} 
MFOL is a fragment of FOL in which all predicates are unary, yet it preserves essential quantified reasoning patterns. Its restricted structure provides greater controllability, making it particularly suitable for scenarios where we aim to systematically manipulate various aspects of logical forms. This enables rich reasoning within individual entities while keeping the structures manageable for dynamic dataset construction.

\noindent \textbf{Task Formulation.}
An MFOL reasoning task~\task{} consists of a set of premises~\premises{}, a candidate conclusion~\conclusion{} and a correctness label
\begin{math}
    \correctnesslabel{} \in \{\true{}, \false{}, \unknown{}\} 
\end{math},
denoted as
\begin{math}
    \task{} = \langle \premises{}, \conclusion{}, \correctnesslabel{} \rangle
\end{math}.
When \correctnesslabel{} is \true{}, \conclusion{} is entailed by \premises{}~(\begin{math}
    \premises{}\models\conclusion{}
\end{math}).
When \correctnesslabel{} is \false{}, \conclusion{} is inconsistent with \premises{}~(\begin{math}
    \premises{}\models\neg\conclusion{}
\end{math}).
When \correctnesslabel{} is \unknown{}, neither \conclusion{} nor $\neg$\conclusion{} is entailed by \premises{}.
The premises~\premises{} consist of a set of rules
\begin{math}
    \rules{} = \{r_1, r_2, \dots, r_m\}
\end{math}
and facts
\begin{math}
    \facts{} = \{f_1, f_2, \dots, f_n\}
\end{math},
denoted as
\begin{math}
    \premises{} = \langle \rules{}, \facts{} \rangle
\end{math}.
In our tasks, each rule $r_i~(1 \le i\le m)$ and fact $f_j~(1 \le j \le n)$ correspond to an MFOL formula.

\noindent \textbf{Minimal Rule.}
Consider the formula
\begin{math}
    \psi(x) = \forall x, (\psi_l(x) \rightarrow \psi_r(x))
\end{math}.
Since only unary predicates are involved and based on the same variable, we abbreviate it as $\psi = \psi_l \rightarrow \psi_r$.
An atomic predicate~($P$) or its negation~($\lnot P$) is referred to as a predicate node~($N$).
If the left-hand side (LHS) $\psi_l$ is a conjunction of predicate nodes, the right-hand side (RHS) $\psi_r$ is a disjunction of predicate nodes, and all nodes correspond to distinct atomic predicates, we call $\psi$ a \emph{minimal rule}, written as
\begin{math}
    \psi = N_{l_1} \land \dots \land N_{l_m} \rightarrow N_{r_1} \lor \dots \lor N_{r_n}
\end{math}. 
This definition is justified as follows.
A formula of the form 
\begin{math}
    \psi_l \land (N_{l_{m+1}} \lor N_{l_{m+2}}) \rightarrow \psi_r
\end{math}
can be decomposed into the two formulas
\begin{math}
    \psi_l \land N_{l_{m+1}} \rightarrow \psi_r
\end{math}
and
\begin{math}
    \psi_l \land N_{l_{m+2}} \rightarrow \psi_r
\end{math}.
Thus the original formula is not minimal.
An analogous argument applies to the RHS:
\begin{math}
    \psi_l \rightarrow \psi_r \lor (N_{r_{n+1}} \land N_{r_{n+2}})
\end{math}
is equivalent to
\begin{math}
    \psi_l \rightarrow \psi_r \lor N_{r_{n+1}}
\end{math}
and
\begin{math}
    \psi_l \rightarrow \psi_r \lor N_{r_{n+2}}
\end{math}.

\noindent \textbf{Depth and Width of a Minimal Rule.}
A minimal rule represents a single-step inference from its LHS to its RHS, and thus its depth is 1. 
The width of a minimal rule is defined as the number of logical connectives~($\land, \lor$) it contains. For the rule  $\psi$, the width is $m+n-2$.

\noindent \textbf{Distractor Rule.}
For a premise set $\premises{}$, consider an MFOL rule $r_D$ that shares some predicates with $\premises{}$ but $r_D \notin \premises$. If for any conclusion $\conclusion{}$, the correctness label \correctnesslabel{} derived from $\premises{}$ is the same as that derived from $\premises{} \cup \{r_D\}$, then $r_D$ is called a distractor rule for $\premises{}$.
In other words, $r_D$ is related to $\premises$ by sharing predicates, yet it does not affect the correctness judgment with respect to any conclusion.

\section{Methodology}
\label{sec:methodology}

We propose \toolname{}, an automated framework for generating MFOL reasoning tasks with quantitative control over multiple dimensions. \autoref{fig:overview} illustrates the overall pipeline. \toolname{} consists of two main modules:
\begin{itemize}[leftmargin=15pt]

    \item \textbf{Logic Construction~(\autoref{sec:LogicConstruction}):}
    We construct an initial set of logical rules $\rules{}_{basic}$ based on the conjunctive and disjunctive structure of FOL. Based on $\rules{}_{basic}$, we derive fact-conclusion pairs $\facts{}$ and $\conclusion{}$ with label $\correctnesslabel{}$, forming a basic reasoning task $\task{}_{basic}$. We further introduce distractor rules $\rules{}_{distract}$,
    resulting in an extended task $\task{}_{distract}$ with higher logical difficulty.
    
    \item \textbf{FOL2NL~(\autoref{sec:fol2nl}):}
    Given a formal task $\task{}_{basic}$ or $\task{}_{distract}$, we use an LLM to assign predicate semantics guided by a topic keyword and convert it into natural language. To ensure consistency, the generated text is translated back into FOL by an LLM and verified using an external prover.

\end{itemize}

Given a specified reasoning depth, width, task label, number of distractor rules, and topic keyword, \toolname{} generates the corresponding task. Based on this framework, we construct \benchmarkname{}, a benchmark of 2880 MFOL tasks for systematic evaluation of logical reasoning in LLMs~(\autoref{sec:benchmark}).


\subsection{Logic Construction}
\label{sec:LogicConstruction}


Given specified dimensional parameters, this module generates MFOL reasoning tasks \task{} based on the predefined algorithm.

\vspace{-2mm}
\begin{algorithm}
\small
\caption{Basic Rules Construction}
\label{alg:basic_rules_construction}
\begin{algorithmic}[1]

\Require Depth $D$, Width $W$
\Ensure Returns basic rule set $\rules{}_{basic}$

\Function{BasicRulesConstruct}{$D, W$} 
    \State $\rules{}_{basic} \gets \{~\}$
    \State $N_0 \gets$ \Call{NewPredicateNode}{$1$}
    \For{$i \gets 1~\text{to}~D$}
        \Comment{\commentstyle{Depth Expansion}}
        \State $N_{i} \gets$ \Call{NewPredicateNode}{$1$}
        \State \textsc{Append}($\rules{}_{basic}$, $N_{i-1} \rightarrow N_{i}$)
    \EndFor

    \For{$j \gets D+1~\text{to}~D+W$} 
        \Comment{\commentstyle{Width Expansion}}
        \State $N_{j} \gets$ \Call{NewPredicateNode}{$1$}
        \State $r \gets$ \Call{RandomSelect}{$\rules{}_{basic}$}
        \If{\Call{RandomSelect}{\{$\land$,$\lor$\}} = $\land$}
            \State $r' \gets ((r_{lhs} \land N_j) \rightarrow r_{rhs})$
        \Else
            \State $r' \gets (r_{lhs} \rightarrow (r_{rhs} \lor N_j))$
        \EndIf
        \State \Call{ReplaceRule}{$\rules{}_{basic}, r, r'$} 
    \EndFor
    
    \State \Return $\rules{}_{basic}$ 
    
\EndFunction

\end{algorithmic} 
\end{algorithm}
\vspace{-2mm}

\subsubsection{Basic Task Construction.}
Starting from the initial formula $\forall x, N_0(x)$ with depth 0 and width 0, we expand depth and width separately to form the basic rule set $\rules{}_{basic}$ with depth $D$ and width $W$. Each rule in $\rules_{basic}$ is a minimal rule.
Based on $\rules_{basic}$, we further generate fact-conclusion pairs according to the specified label \correctnesslabel{}, thereby constructing the basic reasoning task $\task{}_{basic}$.

\noindent \textbf{\scalebox{1.2}{\ding{182}} Basic Rules Construction.}
As shown in \autoref{alg:basic_rules_construction},
the construction starts from the initial predicate node $N_0$. 

\begin{itemize}[leftmargin=15pt]

    \item \textbf{Depth Expansion:}~(line 4-6)
    We first introduce $D$ new predicate nodes and construct the corresponding initial rules, completing the depth expansion.
    
    \item \textbf{Width Expansion:}~(line 7-14)
    we perform $W$ rounds of width expansion. In each round, a rule is randomly selected from the current set. A new predicate node is added either to its LHS via conjunction or to its RHS via disjunction. The updated rule replaces the original one, completing a width-expansion step.

\end{itemize}

After all depth and width expansions, the procedure outputs the basic rule set $R_{\mathrm{basic}}$.

\noindent \textbf{\scalebox{1.2}{\ding{183}} Implicit Rule.}
As shown in the upper-left of \autoref{fig:overview}, $\rules{}_{basic}$ can be viewed as a directed acyclic graph~(DAG). Each predicate node corresponds to a node in the graph. The predecessors of a node are combined by conjunction, while the successors are combined by disjunction. Each rule introduces a directed edge from its LHS to its RHS.
Let $\nodes{}_{in}$ be the set of nodes with zero in-degree, and let $\nodes{}_{out}$ represent the set of nodes with zero out-degree. 
We can merge all rules in $\rules{}_{basic}$ into a single implicit rule $R_{imp}$, as shown in \autoref{eq:imp_rule}.
The LHS of $R_{imp}$ is the conjunction of all nodes in $\nodes{}_{in}$, and its RHS is the disjunction of all nodes in $\nodes{}_{out}$.
\begin{equation}
\small
\begin{aligned}
R_{imp}~=~\left( \mathop{\land}\limits_{N_i \in \nodes{}_{in}} N_i \right) \rightarrow \left(\mathop{\lor}\limits_{N_j \in \nodes{}_{out}} N_j \right)
\end{aligned}
\label{eq:imp_rule}
\end{equation}

\noindent \textbf{\scalebox{1.2}{\ding{184}} Fact-Conclusion Pairs Generation.}
Deriving $R_{imp}$ requires traversing all rules in $\rules{}_{basic}$. We use this implicit rule to to generate fact–conclusion pairs associated with different labels.

\begin{itemize}[leftmargin=15pt]
    \item \textbf{True:}
    Given a fact $f$ as defined in \autoref{eq:true_task}, $R_{imp}$ can be reduced to $R_{imp}^f$, where $\nodes{}_{in}^f \subset \nodes{}_{in}$ and $\nodes{}_{out}^f \subset \nodes{}_{out}$.
    We then generate the fact set $\facts{}=\{f\}$ and conclusion $\conclusion{}_T=R_{imp}^f$ for instances labeled \true{}. 
\end{itemize}    
\begin{equation}
\small
\begin{aligned}
f&{ = } \left( \mathop{\land}\limits_{N_i \in \nodes{}_{in}^f} N_i \right) \land \left(\mathop{\land}\limits_{N_j \in \nodes{}_{out}^f} \neg N_j \right) \\
R_{imp}^f&{ = }  \left( \mathop{\land}\limits_{N_i \in \nodes{}_{in}\setminus\nodes{}_{in}^f} N_i \right) \rightarrow \left(\mathop{\lor}\limits_{N_j \in \nodes{}_{out}\setminus\nodes{}_{out}^f} N_j \right)
\end{aligned}
\label{eq:true_task}
\end{equation}
\begin{itemize}[leftmargin=15pt]
    \item \textbf{False:}
    Since $\neg(\psi_l \rightarrow \psi_r) \equiv (\psi_l \land \neg \psi_r)$, $\neg R_{imp}^f$ is equivalent to $\conclusion{}_F$ in \autoref{eq:false_task}. Thus, for the same fact set $\facts = \{f\}$, we generate instances labeled \false{} with conclusion $\conclusion{}_F$.
\end{itemize}
\begin{equation}
\small
\begin{aligned}
\conclusion{}_F~=\left( \mathop{\land}\limits_{N_i \in \nodes{}_{in}\setminus\nodes{}_{in}^f} N_i \right) \land \left(\mathop{\land}\limits_{N_j \in \nodes{}_{out}\setminus\nodes{}_{out}^f} \neg N_j \right)
\end{aligned}
\label{eq:false_task}
\end{equation}

\noindent To generate instances labeled \unknown{}, we employ two strategies. 

\begin{itemize}[leftmargin=15pt]
    \item \textbf{Unknown$_1$:}
    The first strategy applies when the LHS of $R_{imp}^f$ is \true{} and the RHS is a disjunction, making the correct disjunct indeterminate.
    As shown in \autoref{eq:unknown1_task}, $f_{U_1}$ denotes the LHS of $R_{imp}^f$, and $\conclusion_{U_1}$ is a disjunct from its RHS, where $\nodes_{U_1} \subset \nodes_{\mathrm{out}} \setminus \nodes_{\mathrm{out}}^f$. Given $\facts{} = \{f, f_{U_1}\}$, we set the conclusion $\conclusion{}_{U_1}$.

    \item \textbf{Unknown$_2$:}
    The second strategy applies when the RHS of $R_{imp}^f$ is \true{}, but the LHS is uncertain. This case is symmetric to the first. By setting $f_{U_2} = \conclusion_{U_1}$ and $\conclusion{}_{U_2} = f_{U_1}$, we assign the fact set $\facts = \{f, f_{U_2}\}$ and the conclusion $\conclusion_{U_2}$.

\end{itemize}
\begin{equation}
\small
\begin{aligned}
f_{U_1}{ = } \left( \mathop{\land}\limits_{N_i \in \nodes{}_{in}\setminus\nodes{}_{in}^f} N_i \right) ,~\conclusion{}_{U_1}{ = } \left(\mathop{\lor}\limits_{N_j \in \nodes{}_{U_1}} N_j \right)
\end{aligned}
\label{eq:unknown1_task}
\end{equation}

These resulting fact–conclusion pairs can also be instantiated over specific object constants.
After constructing the basic rule set and generating the fact–conclusion pairs, we obtain the basic task
\begin{math}
    \task{}_{basic} = \langle \premises{}_{basic}, \conclusion{}, \correctnesslabel{} \rangle
\end{math}, where $\premises{}_{basic}=\langle \rules{}_{basic},\facts{} \rangle$.

\subsubsection{Distracted Task Construction.}
To increase task difficulty, we generate a distractor rule set $\rules_{distract}$ from $\rules_{basic}$ and augment $\task_{basic}$ with these rules to obtain the distracted task $\task_{distract}$.

\vspace{-2mm}
\begin{algorithm}
\small
\caption{Distractor Rule Generation}
\label{alg:adding_distractor_rules}
\begin{algorithmic}[1]

\Require Basic rule set $\rules{}_{basic}$
\Ensure Returns distractor rule $r'$

\Function{distractorGen}{$\rules{}_{basic}$}
    \If{\Call{RandomSelect}{\{$1,2$\}} = $1$}
        \Comment{\commentstyle{distractor for Rule}}
        \State $r \gets \Call{RandomSelect}{\rules{}_{basic}}$
        \State $\nodes{} \gets \Call{GetAllNodes}{r}$
        \State $N_{old} \gets \Call{RandomSelect}{\nodes{}}$
        \State $N_{new} \gets \Call{NewPredicateNode}{1}$
        \State $r' \gets \Call{ReplaceNode}{r,N_{old},N_{new}}$
    \Else
        \Comment{\commentstyle{distractor for Predicate Node}}
        \State $\nodes{} \gets \Call{GetAllNodes}{\rules{}}$
        \State $N_{selected} \gets \Call{RandomSelect}{\nodes{}} $
        \State $n \gets \Call{RandomInt}{1,4}$
        \State $\nodes{}' \gets \Call{NewPredicateNode}{n}$
        \State $r' \gets \Call{NewRule}{N_{selected},\nodes{}'}$
    \EndIf
    
    \State \Return $r'$
\EndFunction

\end{algorithmic} 
\end{algorithm}
\vspace{-2mm}

\noindent \textbf{\scalebox{1.2}{\ding{185}} Distractor Rules Construction. }
\label{sec:distractor}
For the basic task $\task{}_{basic}$, the target model only needs to integrate all information from $\premises{}_{basic}$ to determine the correctness of the conclusion. 
To further assess model robustness, we introduce distractor rules that are related to the basic rules but do not alter the original inference path.
We employ two distractor strategies. For each distractor rule, one strategy is selected at random, as summarized in \autoref{alg:adding_distractor_rules}.

\begin{itemize}[leftmargin=15pt]

    \item \textbf{Distractor for Rule:}~(lines 3-7)
    A rule is randomly selected from $\rules{}_{basic}$, and one of its predicate nodes is replaced with a new predicate to form a distractor rule.
    
    \item \textbf{Distractor for Predicate:}~(lines 9–13)
    Let $\nodes{}_{basic}$ be the set of predicate nodes in $\rules{}_{basic}$. We randomly select one node from $\nodes{}_{basic}$, introduce 1–4 new predicate nodes, and construct a new rule that includes both the selected node and the newly introduced nodes. Unlike the basic rules, this rule is not required to be minimal and can be either a \texttt{forall} or \texttt{exists} rule.

\end{itemize}

By augmenting the basic rule set $\rules_{\mathrm{basic}}$ with the distractor rule set $\rules_{\mathrm{distract}}$, we obtain the distracted tasks $\task{}_{distract}$.

\subsection{FOL2NL}
\label{sec:fol2nl}


Given a specified topic keyword, this module converts the MFOL reasoning tasks generated in \autoref{sec:LogicConstruction} into natural language (NL), yielding corresponding NL reasoning tasks. 

\subsubsection{Translating First-Order Logic into Natural Language.}
We employ state-of-the-art LLMs with prompt engineering to translate FOL statements into NL. Specifically, given a specified topic keyword, the LLM converts the premise set $\premises{}$ and the conclusion $\conclusion{}$ into topic-relevant NL statements. The instruction used for this transformation is illustrated in the middle part of \autoref{fig:overview}, while the detailed prompt structure is detailed in \autoref{fig:fol2nl_prompt}.
The prompt consists of four components: \emph{task definition}, \emph{conversion rules}, \emph{example demonstration}, and \emph{input instance}.

\noindent \textbf{Task definition} specifies a three-step conversion process. 
First, given a topic keyword, the LLM assigns a unified  entity domain to all variables in the input FOL formulas. For example, under the topic of \emph{University}, all variables may be instantiated as students within a university.
Second, the LLM assigns topic-relevant semantics to all predicates while preserving the given logical structure. All predicates are positive and mutually independent, preventing conflicts, overlaps, or implicit entailments beyond the FOL formulas. 
Finally, each FOL formula is translated into a natural language sentence based on the assigned predicate semantics.

\noindent \textbf{Conversion rules} specify how FOL symbols are mapped to their corresponding natural language expressions.

\begin{figure}[t!]
\resizebox{0.48\textwidth}{!}{
\begin{tcolorbox}[enhanced, boxrule=1pt,
title=Prompt Template for FOL2NL Conversion,
left=2pt, right=2pt, top=2pt, bottom=2pt, boxsep=2pt,fontupper=\scriptsize]

\textcolor[HTML]{4169E1}{\textbf{\#\#\# Task}} \\
1) Choose a coherent domain of discourse for x that fits the topic. \\
2) Make a Mapping List: \\
- Assign a clear, affirmative (non-negated) meaning to every predicate.\\
- All predicate meanings must be mutually independent \dots \\
3) Translate each FOL formula into English.

\textcolor[HTML]{4169E1}{\textbf{\#\#\# Conversion Rules}} \\
$\forall x~\phi(x)$: ``For every <entity>, \dots'' or ``Every <entity> \dots''\\
$\exists x~\phi(x)$: ``Exist a/an <entity>, \dots'' \dots

\textcolor[HTML]{4169E1}{\textbf{\#\#\# Example}} \\
Topic: University.\\
FOLs: \\
F1: $\forall$x((P3(x)$\land$P2(x))$\rightarrow$P1(x)). F2: $\forall$x(P1(x)$\rightarrow$(P5(x)$\lor$P1(x))) \dots\\
\#\#\# \\
1) Domain: students at a university.\\
2) Mapping List: \\
P1(x): x is eligible to graduate. P2(x): x is enrolled in the capstone course. P3(x): x has passed the comprehensive exam. \dots \\
3) Natural Language Restatement:\\
S1: For every student, if the student has passed the comprehensive exam and the student is enrolled in the capstone course, then the student is eligible to graduate. S2: \dots 

\textcolor[HTML]{4169E1}{\textbf{\#\#\# Now Do The Task}} \\
Process the following inputs without additional explanation output.\\
Topic: \textcolor{blue}{[TOPIC]}\\
FOLs: \textcolor{blue}{[FOLs]}

\end{tcolorbox}
}
\vspace{-0.6cm}
\caption{Prompt Template for FOL2NL conversion.}
\vspace{-0.6cm}
    \label{fig:fol2nl_prompt}
\end{figure}

\noindent \textbf{Example demonstration} illustrates the input–output format of the conversion process. 
The input includes consists of a topic keyword and a set of FOL formulas. The output aligns with the three stages of the task definition: 1) the variable domain, 2) semantic assignments for all predicates, and 3) the NL sentences generated from each FOL formula. Both the FOL formulas and the generated NL sentences are explicitly indexed to ensure one-to-one correspondence and avoid missing or misaligned conversions.

\noindent \textbf{Input instance} includes the specified topic and all FOL formulas to be converted. For each task, the conversion is performed in a single pass, covering the basic rules $\rules{}_{basic}$, distractor rules $\rules{}_{distract}$, facts \facts{}, and the conclusion \conclusion{}.

\subsubsection{Logical Consistency Verification.}
When converting MFOL tasks into NL using an LLM, processing multiple statements simultaneously inevitably introduces translation errors. Manual verification is impractical; therefore, we adopt automatic validation by translating the generated NL back into FOL and verifying it with an external automated theorem prover (ATP)~\cite{atp}.

\noindent \textbf{Impact of Translation Errors.}
While some translation errors are negligible, others may alter the logical relationship between the premises \premises{} and the conclusion \conclusion{}. 
For instance, due to translation errors, an MFOL task labeled as $\correctnesslabel{}=\true{}$ may instead be interpreted as \false{} or \unknown{} in NL.
Such discrepancies can lead to incorrect ground-truth labels and mislead model evaluation.

\noindent \textbf{NL2FOL.} 
The conversion procedure mirrors the FOL2NL stage. The LLM first abstracts NL semantics into predicates and constructs a mapping, and then converts the NL into FOL based on this mapping. The prompt structure is similar to in \autoref{fig:fol2nl_prompt}, with further details provided in~\cite{QFOL}.

\noindent \textbf{Verification with ATP.} 
An ATP can determine the label of a formalized conclusion \conclusion{} based on formalized premises \premises{}. We input \premises{} and \conclusion{} from the reconstructed FOL into the ATP. 
If the predicted label matches the ground-truth label \label{}, the task passes verification, indicating that the FOL2NL and NL2FOL processes preserve the underlying logical structure. Otherwise, the task is deemed invalid.

During task construction with \toolname{}, only tasks that pass logical consistency verification are retained. Failed tasks are returned to the FOL2NL stage for regeneration, while those exceeding the maximum number of attempts are discarded.

\vspace{-2mm}
\subsection{Benchmark}
\label{sec:benchmark}

Using \toolname{}, we construct \benchmarkname{}, a benchmark containing 2880 MFOL reasoning tasks. The detailed dimensional distribution is shown in \autoref{fig:benchmark_distribution}. 

\benchmarkname{} can be partitioned into subsets along specific dimensions, with all other dimensions controlled to remain consistent within each subset, enabling evaluation of model performance across different dimensions.
Both width and depth take values from $\{5, 10, 15, 20\}$. 
Accordingly, \benchmarkname{} can be partitioned into four subsets by depth and four subsets by width, each comprising 720 tasks. 
By jointly considering both dimensions, the dataset can be further divided into 16 subsets with different combinations of depth and width, each containing 180 tasks.
The number of distractors takes values from $\{0, 5, 10, 15, 20\}$, yielding five subsets of 576 tasks each. A value of 0 denotes base tasks, with each subsequent level adding five distractors.
The non-contiguous selection of values is intended not only to capture performance trends across increasing dimensions but also to extend the upper bound of task difficulty.
The label space comprises \true{}, \false{}, and \unknown{}, enabling partitioning into three label-based subsets, each containing 960 tasks.
\benchmarkname{} includes four topics: two general domains, $Food$ and $Animal$, and two specialized domains, $University$ and $Mathematics$, enabling partition into four subsets of 720 tasks each.

For each task in \benchmarkname{}, the target model is required to assign a \true{}, \false{}, or \unknown{} label to a candidate conclusion \conclusion{} based on the provided premises \premises{}. The evaluation instruction is shown in the lower-right of \autoref{fig:overview}, , with the prompt detailed in~\cite{QFOL}. To avoid reliance on superficial ordering cues, the premises are randomly shuffled before presentation.

\begin{figure*}[t]
    \centering
    \includegraphics[width=0.95\textwidth]{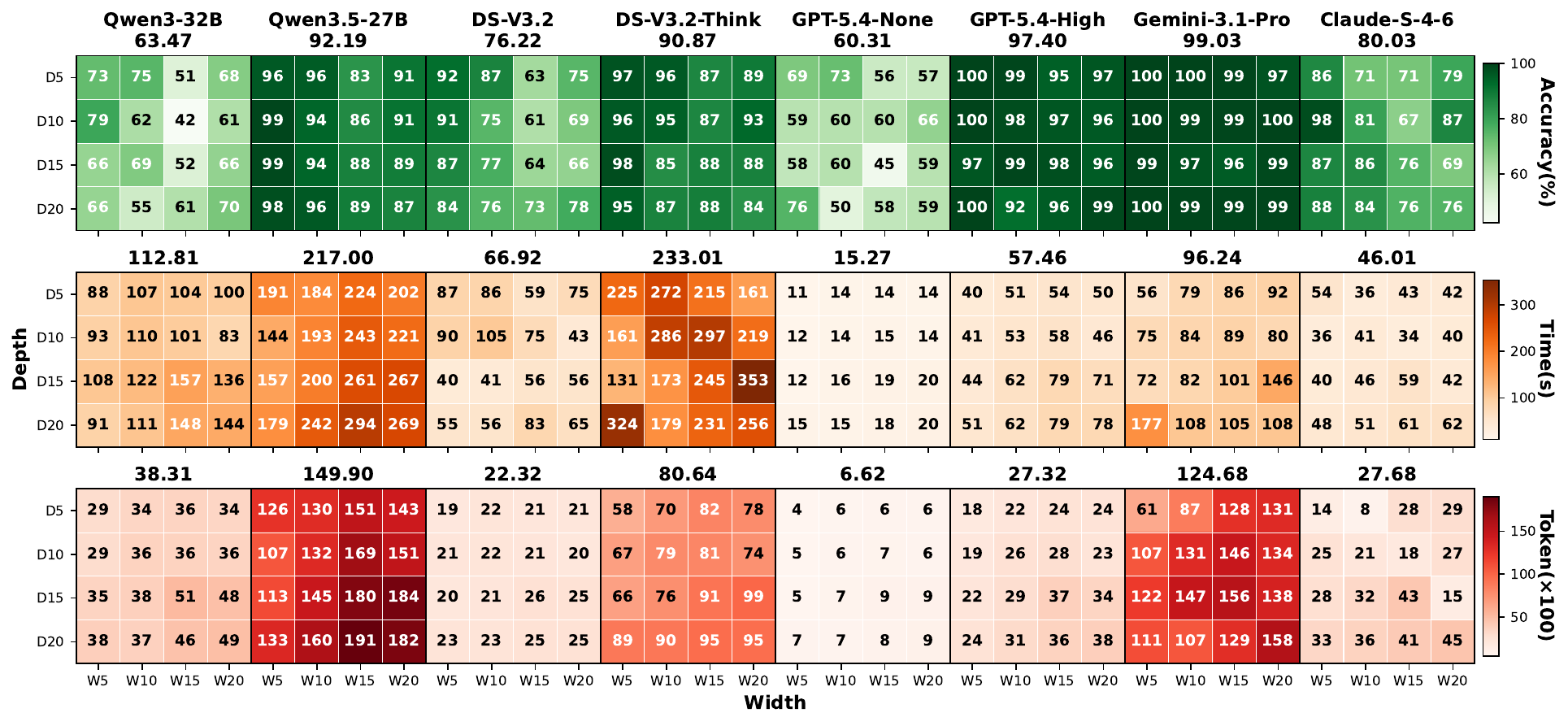}\vspace{-0.4cm}
    \caption{
    Average answer accuracy~(\%), time overhead~(s), and token overhead~($\times$100) on \benchmarkname{} across the overall benchmark and various depth-width subsets. Each 4$\times$4 block represents the overall benchmark with 2880 tasks, with model performance shown at the top. Each small subblock represents a 1/16 subset of the corresponding depth and width configuration, covering 180 tasks. D\emph{$n$}/W\emph{$n$} represents depth/width equal to \emph{$n$}, where \emph{$n\in\{5,10,15,20\}$}.
    }\vspace{-0.3cm}
    \label{fig:rq1}
\end{figure*}

\vspace{-1mm}
\section{Evaluation}
\label{sec:evaluation}
We evaluate the logical reasoning performance of mainstream models on \benchmarkname{} from multiple perspectives, including logical complexity, robustness to distractors, and semantic dependence.
Our evaluation addresses the following research questions:
\begin{itemize}[leftmargin=15pt]
    \item \textbf{RQ1~(Logical Reasoning Capability): }
    How do models perform on \benchmarkname{}, and how does their performance vary with logical factors such as depth, width, and conclusion label?
    \item \textbf{RQ2~(Robustness of Reasoning): } 
    How robust are models to distractor rules?
    \item \textbf{RQ3~(Semantic Dependence): }
    Does topic-specific semantic content affect models’ reasoning under identical logical structures?

\end{itemize}

\noindent \textbf{Evaluation Metrics.} 
We evaluate model performance using single-response answer accuracy, along with computational overheads such as time and token usage.
\benchmarkname{} includes three labels: \true{}, \false{}, and \unknown{}, making it a three-class classification task.
we report label-wise F1 scores (True-F1, False-F1, and Unknown-F1), treating the remaining labels as the ``negative'' class.
We also report Macro-F1, defined as the average F1 score across all classes, giving equal weight to each class regardless of frequency.

\noindent \textbf{Models Under Test.} 
We evaluate six LRMs, including three open-source models, Qwen3-32B~\cite{yang2025qwen3}, Qwen3.5-27B~\cite{qwen35}, and DeepSeek-V3.2-Thinking~\cite{deepseekai2025deepseekv32}, as well as three state of the art closed-source models, Gemini-3.1-Pro-Preview~\cite{gemini31pro}, Claude-Sonnet-4-6\footnote{These are shortened as DS-V3.2-Think, Gemini-3.1-Pro, and Claude-S-4-6, respectively.}~\cite{claudesonnet46}, and GPT-5.4-High\footnote{GPT-5.4-High refers to GPT-5.4 with high $reasoning\_effort$.}~\cite{gpt54}.
For comparison, we further include two LLMs without reasoning modes, DeepSeek-V3.2 and GPT-5.4-None. 

\noindent \textbf{Model Configuration.} 
For reasoning configuration, the Qwen and DeepSeek series only support enabling or disabling \textit{reasoning} mode.
In contrast, GPT, Gemini, and Claude allow control over reasoning intensity, and we consistently set \textit{reasoning\_effort} to high for these models.
For decoding parameters, we follow the official recommended settings, for Qwen3-32B\footnote{\url{https://www.modelscope.cn/models/Qwen/Qwen3-32B}} and Qwen3.5-27B\footnote{\url{https://www.modelscope.cn/models/Qwen/Qwen3.5-27B}} when performing reasoning tasks. The DeepSeek series does not support parameter tuning. 
For GPT, Gemini, and Claude, we retain the default configurations and set the temperature to 1, as lower values may lead to unexpected behavior, such as looping or degraded performance, particularly on complex reasoning tasks.

\noindent \textbf{Implementation.} 
We use DS-V3.2-Think as the translation model for both the FOL2NL stage and the NL2FOL stage. The automated theorem prover used for verification is Vampire~\cite{kovacs2013vampire}.

\noindent \textbf{Running Environment.} All experiments run on a server with Ubuntu 22.04, equipped with two 64-core AMD EPYC 7713 CPUs, 512 GB RAM, and two NVIDIA A100 PCIe 80GB GPUs.

\subsection{RQ1: Main Results}
\label{sec:main_res}
We evaluate model performance on \benchmarkname{}, analyzing both overall performance and its variation across different logical structures (RQ1), including depth, width, and conclusion labels. Additionally, we conduct sample-level analysis to identify and summarize the causes of errors in model responses.

\begin{table}[!htbp]
\centering
\fontsize{9}{11}\selectfont
\caption{
Macro, True, False, and Unknown F1 scores~(\%) for model answers on \benchmarkname{}, with label-wise F1 gains and drops relative to Macro-F1 shown in \textcolor{mRed}{red} and \textcolor{mGreen}{green}.
}
\vspace{-0.3cm}
\label{table:rq1_f1}
\resizebox{0.48\textwidth}{!}{
\begin{tabular}{c|>{\centering\arraybackslash}p{0.8cm}>{\centering\arraybackslash}p{1.1cm}>{\centering\arraybackslash}p{1.1cm}>{\centering\arraybackslash}p{1.4cm}
}
\Xhline{1pt}
\multicolumn{1}{c|}{\multirow{2}{*} {
\diagbox[width=7.5em,height=2.3em]{\textbf{Model}}{\textbf{Metrix}}}}   
& \multicolumn{1}{c}{\multirow{2}{*} {\textbf{Macro}}}   
& \multicolumn{1}{c}{\multirow{2}{*} {\textbf{True}}}
& \multicolumn{1}{c}{\multirow{2}{*} {\textbf{False}}}
& \multicolumn{1}{c}{\multirow{2}{*} {\textbf{Unknown}}}
\\
\multicolumn{1}{c|}{}      & \multicolumn{1}{c}{}   &\multicolumn{1}{c}{} & \multicolumn{1}{c}{} & \multicolumn{1}{c}{}\\
\hline
\textbf{Qwen3-32B} & 
63.29 & \arrowcomp{63.29}{68.89}{5.61} & \arrowcomp{63.29}{58.89}{4.40} & \arrowcomp{63.29}{62.08}{1.21}
\\
\textbf{Qwen3.5-27B} & 
92.19 & \arrowcomp{92.19}{95.39}{3.20} & \arrowcomp{92.19}{92.21}{0.02} & \arrowcomp{92.19}{88.98}{3.22}
\\
\textbf{DS-V3.2} & 
76.27 & \arrowcomp{76.27}{83.02}{6.75} & \arrowcomp{76.27}{74.43}{1.84} & \arrowcomp{76.27}{71.36}{4.91}
\\
\textbf{DS-V3.2-Think} & 
90.77 & \arrowcomp{90.77}{94.21}{3.44} & \arrowcomp{90.77}{89.71}{1.06} & \arrowcomp{90.77}{88.38}{2.38}
\\
\textbf{GPT-5.4-None} & 
55.08 & \arrowcomp{55.08}{83.29}{28.21} & \arrowcomp{55.08}{21.11}{33.97} & \arrowcomp{55.08}{60.83}{5.76}
\\
\textbf{GPT-5.4-High} & 
97.40 & \arrowcomp{97.40}{99.07}{1.67} & \arrowcomp{97.40}{96.73}{0.67} & \arrowcomp{97.40}{96.40}{1.00}
\\
\textbf{Gemini-3.1-Pro} & 
99.03 & \arrowcomp{99.03}{99.22}{0.20} & \arrowcomp{99.03}{99.32}{0.30} & \arrowcomp{99.03}{98.53}{0.50}
\\
\textbf{Claude-S-4-6} & 
79.70 & \arrowcomp{79.70}{92.66}{12.96} & \arrowcomp{79.70}{74.18}{5.52} & \arrowcomp{79.70}{72.25}{7.45}
\\
\Xhline{1pt}
\end{tabular}}
\vspace{-5mm}
\end{table}

\subsubsection{Overall Performance.}
The overall performance of the models on \benchmarkname{} is shown in \autoref{fig:rq1}, with data above each 4$\times$4 block. 
Gemini-3.1-Pro and GPT-5.4-High lead with accuracies of 99.03\% and 97.40\%, respectively, followed by Qwen3.5-27B and DS-V3.2-Think, both exceeding 90\%. In contrast, Claude-S-4-6 achieves only 80.03\% despite with high $reasoning\_effort$. Even with reasoning disabled, DS-V3.2 attains 76.22\%, outperforming GPT-5.4-None at 60.31\%. Despite having more parameters, Qwen3-32B underperforms Qwen3.5-27B, achieving only 63.47\%. 
In terms of overheads, DS-V3.2-Think and Qwen3.5-27B have the highest time costs, averaging 233.01s and 217.00s per task, followed by Qwen3-32B and Gemini-3.1-Pro. For token usage, Qwen3.5-27B and Gemini-3.1-Pro incur the highest costs, with 14990 and 12468 tokens per task, respectively, while DS-V3.2-Think comes next. 
Although Gemini-3.1-Pro, DS-V3.2-Think, and Qwen3.5-27B offer strong performance, their overheads are relatively high. However, Qwen3.5-27B, being a smaller open-source model, still performs admirably. 
Overall, GPT-5.4-High offers the best performance–efficiency trade-off, with an average time of 57.46s and token usage of 2732.

\begin{table}[!htbp]
\centering
\fontsize{9}{11}\selectfont
\caption{
Model answer distribution~(\%) on \benchmarkname{} subsets across different ground truth label.
}
\vspace{-0.3cm}
\label{table:rq1_label_distribution}
\resizebox{0.48\textwidth}{!}{
\begin{tabular}{>{\centering\arraybackslash}p{2.25cm}|>{\centering\arraybackslash}p{0.5cm}>{\centering\arraybackslash}p{0.5cm}>{\centering\arraybackslash}p{0.5cm}|>{\centering\arraybackslash}p{0.5cm}>{\centering\arraybackslash}p{0.5cm}>{\centering\arraybackslash}p{0.5cm}|>{\centering\arraybackslash}p{0.5cm}>{\centering\arraybackslash}p{0.5cm}>{\centering\arraybackslash}p{0.5cm}}
\Xhline{1pt}
\textbf{Groud Truth} & 
\multicolumn{3}{c|}{\multirow{1}{*} {\textbf{True~(T)}}} & \multicolumn{3}{c|}{\multirow{1}{*} {\textbf{False~(F)}}} &
\multicolumn{3}{c}{\multirow{1}{*} {\textbf{Unknwon~(U)}}} \\
\textbf{Model Answer} & 
\textbf{T} & \textbf{F} & \textbf{U} &
\textbf{T} & \textbf{F} & \textbf{U} &
\textbf{T} & \textbf{F} & \textbf{U} \\
\hline
\textbf{Qwen3-32B}
 & \cellcolor{mGreen!70} 70.9
 & \cellcolor{mGreen!6} 6.9
 & \cellcolor{mGreen!22} 22.2
 & \cellcolor{mRed!20} 20.6
 & \cellcolor{mRed!52} 52.3
 & \cellcolor{mRed!27} 27.1
 & \cellcolor{mOrange!14} 14.4
 & \cellcolor{mOrange!18} 18.4
 & \cellcolor{mOrange!67} 67.2
\\
\textbf{Qwen3.5-27B}
 & \cellcolor{mGreen!98} 98.1
 & \cellcolor{mGreen!0} 0.0
 & \cellcolor{mGreen!1} 1.9
 & \cellcolor{mRed!0} 0.6
 & \cellcolor{mRed!86} 86.4
 & \cellcolor{mRed!13} 13.0
 & \cellcolor{mOrange!6} 7.0
 & \cellcolor{mOrange!0} 0.9
 & \cellcolor{mOrange!92} 92.1
\\
\textbf{DS-V3.2}
 & \cellcolor{mGreen!83} 83.5
 & \cellcolor{mGreen!3} 3.9
 & \cellcolor{mGreen!12} 12.6
 & \cellcolor{mRed!6} 6.0
 & \cellcolor{mRed!68} 68.5
 & \cellcolor{mRed!25} 25.4
 & \cellcolor{mOrange!11} 11.7
 & \cellcolor{mOrange!11} 11.8
 & \cellcolor{mOrange!76} 76.6
\\
\textbf{DS-V3.2-Think}
 & \cellcolor{mGreen!99} 99.1
 & \cellcolor{mGreen!0} 0.3
 & \cellcolor{mGreen!0} 0.6
 & \cellcolor{mRed!5} 5.2
 & \cellcolor{mRed!90} 90.3
 & \cellcolor{mRed!4} 4.5
 & \cellcolor{mOrange!6} 6.0
 & \cellcolor{mOrange!10} 10.7
 & \cellcolor{mOrange!83} 83.2
\\
\textbf{GPT-5.4-None}
 & \cellcolor{mGreen!77} 77.1
 & \cellcolor{mGreen!0} 0.2
 & \cellcolor{mGreen!22} 22.7
 & \cellcolor{mRed!0} 0.4
 & \cellcolor{mRed!11} 11.9
 & \cellcolor{mRed!87} 87.7
 & \cellcolor{mOrange!7} 7.6
 & \cellcolor{mOrange!0} 0.4
 & \cellcolor{mOrange!91} 92.0
\\
\textbf{GPT-5.4-High}
 & \cellcolor{mGreen!99} 99.4
 & \cellcolor{mGreen!0} 0.0
 & \cellcolor{mGreen!0} 0.6
 & \cellcolor{mRed!0} 0.4
 & \cellcolor{mRed!93} 93.9
 & \cellcolor{mRed!5} 5.7
 & \cellcolor{mOrange!0} 0.8
 & \cellcolor{mOrange!0} 0.2
 & \cellcolor{mOrange!98} 99.0
\\
\textbf{Gemini-3.1-Pro}
 & \cellcolor{mGreen!100} 100.0
 & \cellcolor{mGreen!0} 0.0
 & \cellcolor{mGreen!0} 0.0
 & \cellcolor{mRed!0} 0.0
 & \cellcolor{mRed!99} 99.3
 & \cellcolor{mRed!0} 0.7
 & \cellcolor{mOrange!1} 1.6
 & \cellcolor{mOrange!0} 0.6
 & \cellcolor{mOrange!97} 97.8
\\
\textbf{Claude-S-4-6}
 & \cellcolor{mGreen!98} 98.0
 & \cellcolor{mGreen!0} 0.1
 & \cellcolor{mGreen!1} 1.9
 & \cellcolor{mRed!2} 2.5
 & \cellcolor{mRed!67} 67.5
 & \cellcolor{mRed!30} 30.0
 & \cellcolor{mOrange!11} 11.0
 & \cellcolor{mOrange!14} 14.4
 & \cellcolor{mOrange!74} 74.6
\\
\Xhline{1pt}
\end{tabular}}
\vspace{-4mm}
\end{table}

\subsubsection{Effect of Depth and Width.}
The model performance across depth–width subsets of \benchmarkname{} is shown in the small blocks of \autoref{fig:rq1}.
When depth and width vary within $\{5, 10, 15, 20\}$, Gemini-3.1-Pro and GPT-5.4-High maintain stable performance, indicating that these tasks remain within their capabilities even at subset with depth and width of 20~(\texttt{D20-W20}), reflecting strong reasoning ability. In contrast, other models exhibit non-linear performance degradation as depth and width increase. Even strong models such as Qwen3.5-27B and DS-V3.2-Think show accuracy drops of 9\% and 13\%, respectively, from the simplest \texttt{D5-W5} subset to the most complex \texttt{D20-W20} subset. Lower-performing models show more pronounced fluctuations, particularly the two LLMs.
Additionally, time and token overheads increase with task difficulty. 
Most high-performing models (e.g., Gemini-3.1-Pro and Qwen3.5-27B) exhibit more pronounced overhead growth than lower-performing ones (e.g., DS-V3.2 and GPT-5.4-None), suggesting deeper reasoning with higher overhead. In contrast, GPT-5.4-High maintains relatively stable overhead.



\begin{table*}[!htbp]
\centering
\caption{
Examples of \textcolor{mGreen}{correct} and \textcolor{mRed}{incorrect} model responses on \benchmarkname{}, with highlights in \textcolor{mblue}{blue} and \textcolor{orange}{orange}.
}\vspace{-0.3cm}
\label{table:rq1_case}
\renewcommand{\arraystretch}{1.2} 
\fontsize{9pt}{12pt}\selectfont
\resizebox{\textwidth}{!}{
\begin{tabular}{p{6.5cm}|p{12.5cm}|p{4.5cm}} 
\Xhline{1pt}
\multicolumn{1}{c|}{\multirow{1}{*}{\textbf{Task Information}}} & 
\multicolumn{1}{c|}{\multirow{1}{*}{\textbf{Model Response}}} & 
\multicolumn{1}{c}{\multirow{1}{*}{\textbf{Description}}}
\\
\hline
\textbf{Task \scalebox{1.1}{\ding{172}}:} 
\textbf{Food\_d5\_w5\_0\_0\_distract15}
\newline
\textbf{Premise 5:} For every food item, if the food item is perishable and the food item is served hot, then the food item is not sour or \dots
\newline
\textbf{Conclusion:}
For every food item, if the food item is served hot, then the food item is not sour.~(\textbf{\true{}})
&
\textbf{(Qwen3-32B)}
\textbf{Explanation:}
\dots
Step 3: \textcolor{orange}{For a non-perishable food item served hot, the premise about ``perishable and served hot'' does not activate.} Therefore, the conclusion ``if served hot, then not sour'' cannot be enforced for non-perishable items, as the premises provide no restriction on their sourness.
Step 4: The conclusion claims a universal rule (``for every food item''), but the premises only restrict sourness in perishable items served hot. \textcolor{orange}{Since non-perishable items are not addressed, the conclusion cannot be guaranteed to hold in all cases.}
\textbf{Answer:} \textcolor{mRed}{\textbf{\unknown{}}}
&
\textbf{Qwen3-32B} fails to leverage the premises to derive ``every food item is perishable'', relying instead on an incorrect assumption ``non\-perishable'' that leads to an erroneous answer.
\\
\hline
\textbf{Task \scalebox{1.1}{\ding{173}}:} 
\textbf{Animal\_d5\_w15\_5\_1\_distract0}
\newline
\textbf{Conclusion:}
There exists an animal such that the animal builds nests and does not have stripes and has webbed feet.~(\textbf{\false{}})
&
\textbf{(DS-V3.2)}
\textbf{Explanation:}
\dots
Step 10: Since for any animal, \textcolor{mblue}{if it lacks stripes and has webbed feet, it cannot build nests}, the conclusion ``builds nests and does not have stripes and has webbed feet'' can never be true for any animal. Therefore the conclusion must be false.
\textbf{Answer:} \textcolor{mGreen}{\textbf{\false{}}}
\newline
\textbf{(GPT-5.4-High)}
\textbf{Explanation:}
\dots
Step 6: So \textcolor{orange}{the premises do not establish the existence of any animal that has all three required properties}: builds nests, does not have stripes, has webbed feet.
Step 7: The conclusion is also not forced to be false, because \textcolor{orange}{the premises never say that no animal can have those properties}.
\textbf{Answer:} \textcolor{mRed}{\textbf{\unknown{}}} 
&
\textbf{DS-V3.2} deduces a contradiction between the premises and conclusion, correctly answering \false{}.
\newline
\textbf{GPT-5.4-High} neither infers the conclusion nor identifies the contradiction, incorrectly marking it as \unknown{}.
\\
\hline
\textbf{Task \scalebox{1.1}{\ding{174}}:} 
\textbf{Mathematics\_d5\_w20\_206\_2\_distract0}
\newline
\textbf{Premise 6:} For every natural number $x$, $x$ is a multiple of 2 \dots
\newline
\textbf{Conclusion:}
For every natural number $x$, $x$ is not a Mersenne number.~(\textbf{\unknown{}})
&
\textbf{(DS-V3.2-Think)}
\textbf{Explanation:}
Step 1: The last premise states that for every natural number $x$, $x$ is a multiple of 2. \textcolor{orange}{This means every natural number is even.}
Step 2: \textcolor{orange}{Mersenne numbers are defined as numbers of the form $2^n-1$ for integer $n\geq1$.} For $n\geq1$, $2^n$ is even, so $2^n - 1$ is odd.
Step 3: Since every natural number is even, no natural number can be odd. \textcolor{orange}{Therefore, no natural number can be a Mersenne number}.
\dots
\textbf{Answer:} \textcolor{mRed}{\textbf{\true{}}}
&
\textbf{DS-V3.2-Think} relies on external knowledge to determine the conclusion, instead of using the logical relationships in the premises.
\\
\Xhline{1pt}
\end{tabular}
}
\vspace{-0.3cm}
\end{table*}

\subsubsection{Effect of Task Label.}
As shown in \autoref{table:rq1_f1}, the various F1 scores indicate that most models perform worse on \false{} and \unknown{} than on \true{}.
Gemini-3.1-Pro and GPT-5.4-High remain stable across labels, whereas GPT-5.4-None and Claude-S-4-6 show the largest fluctuations, with True-F1 exceeding False-F1 by 62.18\% for GPT-5.4-None and exceeding Unknown-F1 by 20.41\% for Claude-S-4-6.
To further analyze label-specific behavior, we examine the response distribution across label subsets, as shown in \autoref{table:rq1_label_distribution}.
Except for Qwen3-32B and the two LLMs, which often misclassify \true{} as \unknown{}, all other models exceed 98\% accuracy on the \true{} subset.
For the \false{} subset, only Gemini-3.1-Pro maintains high accuracy. Even GPT-5.4-High misclassifies 5.7\% of \false{} cases as \unknown{}. Other models also exhibit a strong tendency to predict \unknown{}, particularly GPT-5.4-None.
For the \unknown{} subset, errors are roughly balanced between \true{} and \false{}.

\subsubsection{Model Response Error Analysis.} 
We manually analyze model response errors, with representative examples shown in \autoref{table:rq1_case}.
For tasks labeled \true{}, most errors stem from insufficient reasoning. For example, Qwen3-32B's response on Task \scalebox{1.1}{\ding{172}} fails to reason deeply from the premises, relying instead on an incorrect assumption, which leads to an incorrect answer. 
Similar issues arise for \false{} tasks.
The accuracy on the \false{} subset is generally lower than on \true{}, partly due to the model's forward reasoning habit.
For \true{} tasks, the model can directly infer the conclusion from the premises. However, for \false{} tasks, the model typically identifies contradictions step by step, rather than directly reasoning the inverse of the conclusion, as shown in DS-V3.2's response on Task \scalebox{1.1}{\ding{173}}.
When the model cannot reason the conclusion or identify contradictions, it often responds with \unknown{}, as seen in GPT-5.4-High’s response on Task \scalebox{1.1}{\ding{173}}. This issue ultimately stems from insufficient reasoning.
The accuracy on \unknown{} tasks is low, partly due to content that conflicts with real-world knowledge. When the model cannot reach a definite \false{} or \unknown{} conclusion, it tends to incorporate external knowledge, as shown in DS-V3.2-Think's response on Task \scalebox{1.1}{\ding{174}}.
Although, when handling deductive reasoning tasks, the model should ideally rely solely on the premises.

\begin{tcolorbox}[title=ANSWER to RQ1, boxrule=0.8pt,boxsep=1.5pt,left=2pt,right=2pt,top=2pt,bottom=1pt,fontupper=\small]
Gemini-3.1-Pro achieves the highest accuracy, while GPT-5.4-High offers the best performance-overhead trade-off on \benchmarkname{}. 
As depth and width increase, performance degrades non-linearly and overhead rises. Across labels, models perform better on \true{} than on \false{} and \unknown{}, with errors on \true{} and \false{} tasks often misclassified as \unknown{}.

\end{tcolorbox} 

\begin{figure*}[t]
\vspace{-0.3cm}
    \centering
    \includegraphics[width=0.95\textwidth]{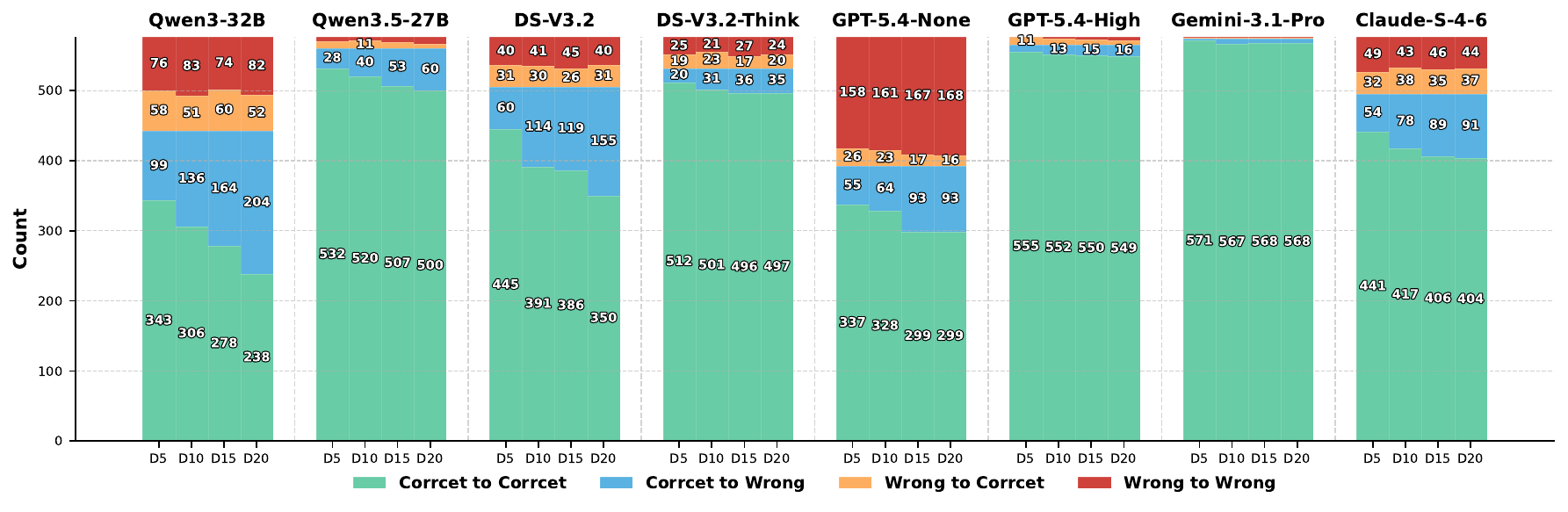}\vspace{-0.4cm}
    \caption{
    Model answer changes from \benchmarkname{} subsets without distractors to those with varying numbers of distractors, each containing 576 tasks. D\emph{$n$} represents a subset with \emph{$n$} distractor rules, where \emph{$n \in \{5,10,15,20\}$}. 
    ``Correct/Wrong to Correct/Wrong'' indicates the model answers correctly/incorrectly in a D0 task and correctly/incorrectly in the corresponding D$n$ task.
    }\vspace{-0.3cm}
    \label{fig:rq2}
\end{figure*}

\subsection{RQ2: Robustness} 
In addition to logical complexity, we evaluate model robustness under increasing numbers of distractor rules (RQ2) and conduct an ablation study to determine whether observed changes are attributable to increased text length.

\begin{table}[!htbp]
\centering
\fontsize{9}{12}\selectfont\vspace{-0.2cm}
\caption{Average answer accuracy~(\%) on \benchmarkname{} subsets across different distractors, each containing 576 tasks. D\emph{$n$} represents \emph{$n$} distractor rules, where \emph{$n\in\{0,5,10,15,20\}$}.
}\vspace{-0.3cm}
\label{table:rq2}
\centering
\resizebox{0.48\textwidth}{!}{
\begin{tabular}{c|ccccc}
\Xhline{1pt}
\multicolumn{1}{c|}{\multirow{2}{*} {\diagbox[width=9em,height=2.6em]{\textbf{Model}}{\textbf{Subset}}}} 
& 
\multicolumn{1}{c}{\multirow{2}{*}{\textbf{D0}}} &
\multicolumn{1}{c}{\multirow{2}{*}{\textbf{D5}}} &
\multicolumn{1}{c}{\multirow{2}{*}{\textbf{D10}}} &
\multicolumn{1}{c}{\multirow{2}{*}{\textbf{D15}}} &
\multicolumn{1}{c}{\multirow{2}{*}{\textbf{D20}}} 
\\
\\
\hline
\multicolumn{1}{c|}{\textbf{Qwen3-32B}} & 
\cellcolor{mGreen!53} 76.74 & \cellcolor{mGreen!39} 69.62 & \cellcolor{mGreen!23} 61.98 & \cellcolor{mGreen!16} 58.68 & \cellcolor{mGreen!0} 50.35 
\\
\multicolumn{1}{c|}{\textbf{Qwen3.5-27B}} & 
\cellcolor{mGreen!95} 97.22 & \cellcolor{mGreen!88} 94.10 & \cellcolor{mGreen!84} 92.19 & \cellcolor{mGreen!79} 89.58 & \cellcolor{mGreen!76} 87.85 
\\
\multicolumn{1}{c|}{\textbf{DS-V3.2}} & 
\cellcolor{mGreen!75} 87.67 & \cellcolor{mGreen!65} 82.64 & \cellcolor{mGreen!46} 73.09 & \cellcolor{mGreen!42} 71.53 & \cellcolor{mGreen!32} 66.15 
\\
\multicolumn{1}{c|}{\textbf{DS-V3.2-Think}} & 
\cellcolor{mGreen!85} 92.36 & \cellcolor{mGreen!84} 92.19 & \cellcolor{mGreen!82} 90.97 & \cellcolor{mGreen!78} 89.06 & \cellcolor{mGreen!79} 89.76 
\\
\multicolumn{1}{c|}{\textbf{GPT-5.4-None}} & 
\cellcolor{mGreen!35} 68.06 & \cellcolor{mGreen!25} 63.02 & \cellcolor{mGreen!21} 60.94 & \cellcolor{mGreen!9} 54.86 & \cellcolor{mGreen!8} 54.69 
\\
\multicolumn{1}{c|}{\textbf{GPT-5.4-High}} & 
\cellcolor{mGreen!96} 98.09 & \cellcolor{mGreen!97} 98.26 & \cellcolor{mGreen!95} 97.40 & \cellcolor{mGreen!94} 96.88 & \cellcolor{mGreen!93} 96.35 
\\
\multicolumn{1}{c|}{\textbf{Gemini-3.1-Pro}} & 
\cellcolor{mGreen!100} 99.65 & \cellcolor{mGreen!99} 99.31 & \cellcolor{mGreen!97} 98.61 & \cellcolor{mGreen!98} 98.78 & \cellcolor{mGreen!98} 98.78 
\\
\multicolumn{1}{c|}{\textbf{Claude-S-4-6}} & 
\cellcolor{mGreen!72} 85.94 & \cellcolor{mGreen!64} 82.12 & \cellcolor{mGreen!58} 78.99 & \cellcolor{mGreen!53} 76.56 & \cellcolor{mGreen!53} 76.56 
\\
\Xhline{1pt}
\end{tabular}}
\vspace{-4mm}
\end{table}

\subsubsection{Effect of Distractor.}
As shown in \autoref{table:rq2}, model accuracy decreases from the \texttt{D0} subset (tasks with no distractors) to the \texttt{D20} subset (tasks with 20 distractors). Gemini-3.1-Pro, GPT-5.4-High, and DS-V3.2-Think show smaller declines, indicating stronger robustness to distractor factors, while other models experience larger drops. Notably, Qwen3-32B shows the largest decline with a 26.39\% performance drop. Qwen3.5-27B drops from 97.22\% on \texttt{D0} to 87.85\% on \texttt{D20}, whereas DS-V3.2-Think only decreases from 92.36\% to 89.76\%. This suggests that while Qwen3.5-27B has better reasoning performance, its robustness is lower than DS-V3.2-Think.
In terms of overhead, a similar trend holds: increasing distractors lead to higher time and token costs, as reported in~\cite{QFOL}.

Additionally, we examine the specific changes in model responses with and without distractors, as shown in \autoref{fig:rq2}. After adding distractors, some models shift from correct to wrong~(CW) answers, such as in Task \scalebox{1.1}{\ding{172}} in \autoref{table:rq1_case}, where Qwen3-32B answers correctly on \texttt{D0} task~($Food\_d5\_w5\_0\_0\_distract0$) but incorrectly on its corresponding \texttt{D15} task~($Food\_d5\_w5\_0\_0\_distract15$). 
There are also cases where models shift from wrong to correct~(WC) answers, as seen in Task \scalebox{1.1}{\ding{173}} in \autoref{table:rq1_case}, where GPT-5.4-High answers correctly on the \texttt{D5} task~($Animal\_d5\_w15\_5\_1\_distract5$) but incorrectly on the corresponding \texttt{D0} task~($Animal\_d5\_w15\_5\_1\_distract0$).
This suggests that while distractors can hinder the model’s reasoning, they can also stimulate deeper thinking. However, overall, CW instances outnumber WC, and as the number of distractors increases, model performance generally declines.

\subsubsection{Ablation Study.} 
We conduct an ablation study to demonstrate that the performance decline in D20 compared to D0 is caused by the distractors rather than the increased context length. 
Since Qwen3-32B exhibits the lowest robustness in \autoref{table:rq2}, we base our experiments on this model. 
Using its tokenizer, we find that \texttt{D20} tasks contain, on average, 603 more tokens than \texttt{D0}.
To isolate the effect of context length, we append 600 tokens of ``\texttt{Unimportant Content}'' as placeholders to \texttt{D0} tasks. Under this setting, Qwen3-32B achieves an accuracy of 75.69\%, nearly matching its \texttt{D0} accuracy of 76.74\%. 
These results indicate that the observed performance degradation is primarily attributable to distractors, with only minimal impact from increased context length, underscoring the significant effect of the designed distractor rules.

\begin{tcolorbox}[title=ANSWER to RQ2, boxrule=0.8pt,boxsep=1.5pt,left=2pt,right=2pt,top=2pt,bottom=1pt,fontupper=\small]
As distractor rules increase, model performance declines, with Gemini-3.1-Pro showing the best robustness. LRMs generally outperform LLMs in robustness, except for Qwen3-32B, which has relatively poor robustness.
\end{tcolorbox} 

\begin{table}[!htbp]
\centering
\fontsize{9}{11}\selectfont
\caption{
Average answer accuracy~(\%) on different topic subsets of \benchmarkname{}, each containing 720 tasks, showing gains and drops relative to the overall average in \textcolor{mRed}{red} and \textcolor{mGreen}{green}. \emph{Univ.} and \emph{Math} stand for \emph{University} and \emph{Mathematics}.
}
\vspace{-0.3cm}
\label{table:rq3}
\begin{tabular}{c|
>{\centering\arraybackslash}p{1.1cm}
>{\centering\arraybackslash}p{1.1cm}
>{\centering\arraybackslash}p{1.1cm}
>{\centering\arraybackslash}p{1.1cm}}
\Xhline{1pt}
\multicolumn{1}{c|}{\multirow{2}{*} {
\diagbox[width=7.5em,height=2.3em]{\textbf{Model}}{\textbf{Subset}}}}   
& \multicolumn{1}{c}{\multirow{2}{*} {\textbf{Food}}}   
& \multicolumn{1}{c}{\multirow{2}{*} {\textbf{Animal}}}
& \multicolumn{1}{c}{\multirow{2}{*} {\textbf{Univ.}}}
& \multicolumn{1}{c}{\multirow{2}{*} {\textbf{Math}}}\\
\multicolumn{1}{c|}{}      & \multicolumn{1}{c}{}   &\multicolumn{1}{c}{} & \multicolumn{1}{c}{} & \multicolumn{1}{c}{}\\
\hline
\textbf{Qwen3-32B} & 
\arrowcomp{63.47}{63.89}{0.42} & \arrowcomp{63.47}{64.44}{0.97} & \arrowcomp{63.47}{65.42}{1.94} & \arrowcomp{63.47}{60.14}{3.33}
\\
\textbf{Qwen3.5-27B} & 
\arrowcomp{92.19}{90.56}{1.63} & \arrowcomp{92.19}{92.22}{0.03} & \arrowcomp{92.19}{92.92}{0.73} & \arrowcomp{92.19}{93.06}{0.87}
\\
\textbf{DS-V3.2} & 
\arrowcomp{76.22}{78.75}{2.53} & \arrowcomp{76.22}{78.89}{2.67} & \arrowcomp{76.22}{79.72}{3.51} & \arrowcomp{76.22}{67.50}{8.72}
\\
\textbf{DS-V3.2-Think} & 
\arrowcomp{90.87}{93.47}{2.60} & \arrowcomp{90.87}{95.42}{4.55} & \arrowcomp{90.87}{96.67}{5.80} & \arrowcomp{90.87}{77.92}{12.95}
\\
\textbf{GPT-5.4-None} & 
\arrowcomp{60.31}{57.50}{2.81} & \arrowcomp{60.31}{59.86}{0.45} & \arrowcomp{60.31}{61.25}{0.94} & \arrowcomp{60.31}{62.64}{2.33}
\\
\textbf{GPT-5.4-High} & 
\arrowcomp{97.40}{97.50}{0.10} & \arrowcomp{97.40}{96.81}{0.59} & \arrowcomp{97.40}{98.19}{0.80} & \arrowcomp{97.40}{97.08}{0.31}
\\
\textbf{Gemini-3.1-Pro} & 
\arrowcomp{99.03}{98.75}{0.28} & \arrowcomp{99.03}{98.75}{0.28} & \arrowcomp{99.03}{99.86}{0.83} & \arrowcomp{99.03}{98.75}{0.28}
\\
\textbf{Claude-S-4-6} & 
\arrowcomp{80.03}{78.61}{1.42} & \arrowcomp{80.03}{77.78}{2.26} & \arrowcomp{80.03}{84.72}{4.69} & \arrowcomp{80.03}{79.03}{1.01}
\\
\Xhline{1pt}
\end{tabular}
\vspace{-4mm}
\end{table}

\subsection{RQ3: Semantic Dependence}

In this section, we examine whether model performance varies across different semantic topics on \benchmarkname{} under a consistent logical structure.

Model accuracy on different topic subsets of \benchmarkname{} is shown in \autoref{table:rq3}. 
Most models achieve higher average accuracy on the $University$ subset than on other topics and the overall benchmark, suggesting greater familiarity with this domain. This trend is particularly evident for Claude-S-4-6, which attains 84.72\% accuracy on $University$ while remaining below 80\% on other topics.
Gemini-3.1-Pro and GPT-5.4-High show minimal variation across the four topics, indicating stable performance without clear preference. 
Note that the evaluation is limited to four predefined topics, and thus these findings may not generalize to broader real-world domains.
In contrast, GPT-5.4-None and Qwen3.5-27B perform worse on the $Food$ subset, while DS-V3.2-Think, DS-V3.2, and Qwen3-32B show notable degradation on the $Mathematics$ subset. 
In particular, DS-V3.2-Think and DS-V3.2 exhibit accuracy drops of 18.75\% and 12.22\%, respectively, compared to their performance on $University$, highlighting a strong topic sensitivity.

To further investigate this anomaly, we analyze the behavior of DS-V3.2-Think and DS-V3.2 on the $Mathematics$ subset. 
DS-V3.2-Think produces incorrect answers on 16.81\% of tasks in $Mathematics$, despite correctly solving tasks with identical logical structures in the other three topics. Similarly, DS-V3.2 exhibits a 14.44\% error rate.
Examination of these cases shows that both two models tend to incorporate external knowledge not specified in the premises when handling mathematical content, leading to deviations from the intended deductive reasoning process.
An illustrative example is DS-V3.2-Think’s response to Task \scalebox{1.1}{\ding{174}} in \autoref{table:rq1_case}.


\begin{tcolorbox}[title=ANSWER to RQ3, boxrule=0.8pt,boxsep=1.5pt,left=2pt,right=2pt,top=2pt,bottom=1pt,fontupper=\small]
Model performance varies with semantic content, even under identical logical structures. Among the four topics in \benchmarkname{}, models are more familiar with the $University$ topic, while the DeepSeek series performs worst on $Mathematics$.
\end{tcolorbox} 
\section{Discussion}

\subsection{Further Analysis}
We provide additional analysis to evaluate the reliability of \benchmarkname{}, focusing on logical consistency between FOL and NL and the impact of context length on model performance.

\noindent \textbf{Logical Consistency between FOL and NL.} 
Despite our verification mechanism, there is a potential failure mode where errors in both FOL2NL and NL2FOL translations cancel out, allowing incorrect instances to pass. To evaluate this risk, we randomly sampled 50 instances for manual inspection and found no inconsistencies. This suggests that our verification pipeline produces reliable tasks in practice.

\noindent \textbf{Impact of Context Length.}
\benchmarkname{} sets depth, width, and distractor up to 20, with higher dimensions resulting in longer contexts. We analyze the prompt token length of input tasks and observe an average of about 860 tokens overall and about 1380 tokens for the highest-dimensional subset. Both are well within the context window limits of the evaluated models, ranging from 32K (Qwen3-32B) to 1M tokens (Claude-S-4-6). These results suggest that performance on our benchmark is primarily constrained by models’ reasoning capabilities rather than their ability to handle long contexts.

\subsection{Threats to Validity}

\noindent \textbf{Formal Representation.}
Our framework currently focuses on MFOL with unary predicates to enable precise control over logical structure construction and task complexity. Nevertheless, the proposed multidimensional-driven logical structure construction and semantic domain mapping framework can be naturally extended to full FOL with higher-arity predicates, enabling richer relational reasoning and more expressive logical compositions.

\noindent \textbf{Task Variations.}
\toolname{} quantifies multiple reasoning dimensions, and \benchmarkname{} spans 960 task configurations across diverse logical and semantic settings. Building on this foundation, future work can further expand the design space by introducing additional reasoning dimensions, more diverse distractor strategies, and broader task configurations, enabling more comprehensive and fine-grained evaluation of model logical reasoning capabilities.

\noindent \textbf{Translation Model Dependence.}
Our framework uses LLMs to translate formal logic into natural language with round-trip verification to ensure consistency, suggesting that stronger reasoning models better maintain fidelity. In preliminary experiments with GPT-4o~\cite{gpt4o} and Qwen3-32B, we observed frequent deviations from predefined predicate mappings, causing semantic inconsistencies and requiring repeated validation and regeneration. In contrast, DS-V3.2-Think performs well in these aspects and is adopted in our implementation. Future work may explore more cost-efficient models with comparable capability as they emerge.

\section{Related Wrok}
\subsection{LLM Reasoning Capability Testing}
As the reasoning capabilities of models improve, effective evaluation becomes increasingly important. Recent studies assess reasoning from multiple perspectives. 
MATH-Perturb~\cite{huang2025mathperturb} and GSM-Symbolic~\cite{mirzadeh2024gsm} create benchmarks for mathematical reasoning by perturbing existing datasets.
Drowzee~\cite{li2024drowzee} constructs benchmark using knowledge from Wikipedia~\cite{Auer2007DBpediaAN} to assess commonsense reasoning.
ChronoSense~\cite{islakoglu2025chronosense} and \cite{wongchamcharoen2025chronology} build temporal reasoning benchmarks based on real-world event timing relationships.
For logical reasoning, 
SmartyPat~\cite{xu2025socrates} integrate external tools such as Prolog~\cite{prolog} with LLMs to automatically generate benchmarks.
Across these studies, a common strategy is to create task-specific benchmarks to prevent data leakage from existing datasets, typically using semi-automated or fully automated methods for scalable and continuous task generation.


\subsection{Deductive Reasoning Benchmarks}
Deductive reasoning benchmarks are widely used to evaluate the logical reasoning capabilities of language models. 
Early benchmarks, such as RobustLR~\cite{sanyal2022robustlr}, PrOntoQA~\cite{saparov2022prontoqa}, PrOntoQA-OOD~\cite{saparov2023prontoqaood},  RuleTaker~\cite{clark2020ruletaker}, and ProofWriter~\cite{tafjord2021proofwriter}, rely on predefined deductive rules expansion and template-based text generation.
While these approaches offer structural controllability, they lack both logical complexity and real-world semantics.
LogicNLI~\cite{tian2021logicnli} and FOLIO~\cite{han2024folio} introduce richer FOL rules and more natural expressions. However, LogicNLI requires manual refinement during sentence conversion, and FOLIO is entirely human-annotated, limiting scalability.
Recent work ProverQA~\cite{qi2025proverqa}, 
combine external tools and predefined rules for logical construction, then use LLMs to map these to natural language. However, these methods suffer from two limitations: they provide insufficient control over the resulting logical structures, and LLM-based transformations fail to guarantee end-to-end logical consistency.
Overall, existing benchmarks struggle to balance controllability, scalability, semantic diversity, and logical consistency in natural language generation, limiting their suitability as long-term evaluation tools for rapidly evolving models.


\section{Conclusion}

We present \toolname, an automated framework for constructing deductive reasoning tasks with fine-grained, multidimensional control over logical complexity. By systematically generating monadic first-order logic structures and translating them into natural language with round-trip verification, the framework ensures both controllability and logical consistency. Building on this approach, we introduce \benchmarkname, a benchmark spanning diverse logical and semantic configurations that enables more precise evaluation of modern language models. Experimental results show that performance degrades as logical complexity increases, exhibits systematic biases across label types, and varies across semantic domains.
Looking ahead, the framework can be extended to full first-order logic with higher-arity predicates, as well as to richer task configurations and reasoning dimensions.
Overall, \toolname provides a scalable foundation for controllable deductive reasoning benchmark construction and facilitates deeper analysis of language model reasoning capabilities.


\section*{Data Availability Statement}
The code, data, and additional information relevant to this study are available at~\cite{QFOL}.

\bibliographystyle{ACM-Reference-Format}
\bibliography{8.ref}





\end{document}